\documentclass[11pt]{article}
\usepackage[hyperref]{ccl2020-en}
\usepackage{times}
\usepackage{url}
\usepackage{latexsym}
\usepackage{graphicx}
\usepackage{amsmath}
\usepackage{amsthm}
\usepackage{booktabs}
\usepackage{algorithm}
\usepackage{algorithmic}
\usepackage{bm}
\usepackage{amsfonts}
\usepackage{fancyhdr}

\title{Towards Causal Explanation Detection with Pyramid \\ Salient-Aware Network}

\author{Xinyu Zuo$^{1,2}$, Yubo Chen$^{1,2}$, Kang Liu$^{1,2}$, Jun Zhao$^{1,2}$ \\
	$^1$National Laboratory of Pattern Recognition, Institute of Automation, \\
	Chinese Academy of Sciences, Beijing, 100190, China \\
	$^2$School of Artificial Intelligence, \\
	University of Chinese Academy of Sciences, Beijing, 100049, China \\
	{\tt \{xinyu.zuo,yubo.chen,kliu,jzhao\}@nlpr.ia.ac.cn}}

\date{}

\begin{document}
	\maketitle
	\begin{abstract}
		Causal explanation analysis (CEA) can assist us to understand the reasons behind daily events, which has been found very helpful for understanding the coherence of messages. In this paper, we focus on \emph{Causal Explanation Detection}, an important subtask of causal explanation analysis, which determines whether a causal explanation exists in one message. We design a \textbf{P}yramid \textbf{S}alient-\textbf{A}ware \textbf{N}etwork (PSAN) to detect causal explanations on messages. PSAN can assist in causal explanation detection via capturing the salient semantics of discourses contained in their keywords with a bottom graph-based word-level salient network. Furthermore, PSAN can modify the dominance of discourses via a top attention-based discourse-level salient network to enhance explanatory semantics of messages. The experiments on the commonly used dataset of CEA shows that the PSAN outperforms the state-of-the-art method by 1.8\% F1 value on the \emph{Causal Explanation Detection} task.
	\end{abstract}
	
	\section{Introduction}
	\label{intro}
	
\begin{figure}[h]
		\centering
		\includegraphics*[clip=true,width=0.65\textwidth,height=0.25\textheight]{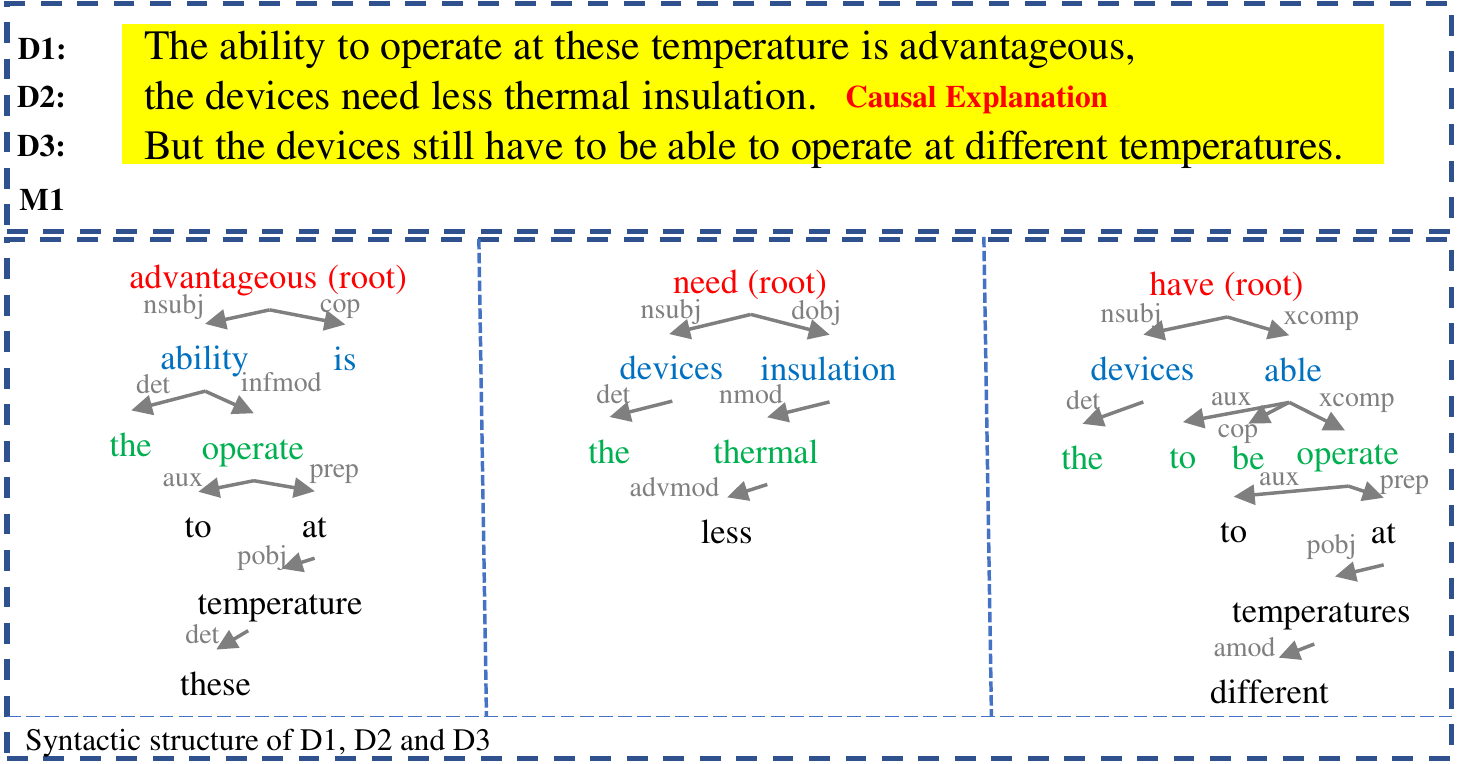}
		\caption{Instance of causal explanation analysis (CEA). The top part is a message which contains its segmented discourses and a causal explanation. The bottom part is the syntactic dependency structures of three discourses divided from M1.} \label{fig1}
	\end{figure}

	Causal explanation detection (CED) aims to detect whether there is a causal explanation in a given message (e.g. a group of sentences). Linguistically, there are coherence relations in messages which explain how the meaning of different textual units can combine to jointly build a discourse meaning for the larger unit. The explanation is an important relation of coherence which refers to the textual unit (e.g. discourse) in a message that expresses explanatory coherent semantics \cite{Jurafsky2010Speech}. As shown in Figure \ref{fig1}, M1 can be divided into three discourses, and D2 is the explanation that expresses the reason why it is advantageous for the equipment to operate at these temperatures. CED is important for tasks that require an understanding of textual expression \cite{son2018causal}. For example, for question answering, the answers of questions are most likely to be in a group of sentences that contains causal explanations \cite{oh2013question}. Furthermore, the summarization of event descriptions can be improved by selecting causally motivated sentences \cite{hidey-mckeown-2016-identifying}. Therefore, CED is a problem worthy of further study.
	
	\label{sec:INTRO}
	
	The existing methods mostly regard this task as a classification problem \cite{son2018causal}. At present, there are mainly two kinds of methods, feature-based methods and neural-based methods, for similar semantic understanding tasks in discourse granularity, such as opinion sentiment classification and discourse parsing \cite{nejat-etal-2017-exploring,jia-etal-2018-modeling,soricut-marcu-2003-sentence}. The feature-based methods can extract the feature of the relation between discourses. However, these methods do not deal well with the implicit instances which lack explicit features. For CED, as shown in Figure \ref{fig1}, D2 lacks explicit features such as \emph{because of}, \emph{due to}, or the features of tenses, which are not friendly for feature-based methods. The methods based on neural network are mainly Tree-LSTM model \cite{wang-etal-2017-tag} and hierarchical Bi-LSTM model \cite{son2018causal}. The Tree-LSTM models learn the relations between words to capture the semantics of discourses more accurately but lack further understanding of the semantics between discourses. The hierarchical Bi-LSTM models can employ sequence structure to implicitly learn the relations between words and discourses. However, previous work shows that compared with Tree-LSTM, Bi-LSTM lacks a direct understanding of the dependency relations between words. Therefore, the method of implicit learning of inter-word relations is not prominent in the tasks related to understanding the semantic relations of messages \cite{li-etal-2015-tree}. Therefore, how to directly learn the relations between words effectively and consider discourse-level correlation to further filter the key information is a valuable point worth studying.
	
	Further analysis, why do the relations between words imply the semantics of the message and its discourses? From the view of computational semantics, the meaning of a text is not only the meaning of words but also the relation, order, and aggregation of the words. In other simple words is that the meaning of a text is partially based on its syntactic structure \cite{Jurafsky2010Speech}. In detail, in CED, the core and subsidiary words of discourses contain their basic semantics. For example, as D1 shown in Figure \ref{fig1}, according to the word order in syntactic structure, we can capture the \emph{ability} of \emph{temperature} is \emph{advantageous}. We can understand the basic semantic of D1 which expresses some kind of \emph{ability} is \emph{advantageous} via root words \emph{advantageous} and its affiliated words. Additionally, why the correlation and key information at the discourse level are so important to capture the causal explanatory semantics of the message? Through observation, the different discourse has a different status for the explanatory semantics of a message. For example, in M1, combined with D1, D2 expresses the explanatory semantics of \emph{why the ability to work at these temperatures is advantageous}, while D3 expresses the semantic of transition. In detail, D1 and D2 are the keys to the explanatory semantics of M1, and if not treated D1, D2, and D3 differently, the transitional semantic of D3 can affect the understanding of the explanatory semantic of M1. Therefore, how to make better use of the information of keywords in the syntactic structure and pay more attention to the discourses that are key to explanatory semantics is a problem to be solved.
	
	To this end, we propose a \textbf{P}yramid \textbf{S}alient-\textbf{A}ware \textbf{N}etworks (PSAN) which utilizes keywords on the syntactic structure of each discourse and focuses on the key discourses that are critical to explanatory semantics to detect causal explanation of messages. First, what are the keywords in a syntactic structure? From the perspective of syntactic dependency, the root word is the central element that dominates other words, while it is not be dominated by any of the other words, all of which are subordinate to the root word \cite{ZongChengqing}. From that, the root and subsidiary words in the dependency structure are the keywords at the syntax level of each discourse. Specifically, we sample 100 positive sentences from training data to illuminate whether the keywords obtained through the syntactic dependency contain the causal explanatory semantics. And we find that the causal explanatory semantics of more than 80\% sentences be captured by keywords in dependency structure\footnote{Five Ph.D. students majoring in NLP judge whether sentences could be identified as which containing causal explanatory semantics by the root word and its surrounding words in syntactic dependency, and the agreement consistency is 0.8}. Therefore, we extract the root word and its surrounding words on the syntactic dependency of each discourse as its keywords. 
	
	Next, we need to consider how to make better use of the information of keywords contained in the syntactic structure. To pay more attention to keywords, the common way is using attention mechanisms to increase the attention weight of them. However, this implicitly learned attention is not very interpretable. Inspired by previous researches \cite{vashishth-etal-2019-incorporating,bastings-etal-2017-graph}, we propose a bottom graph-based word-level salient network which merges the syntactic dependency to capture the salient semantics of discourses contained in their keywords. Finally, how to consider the correlation at the discourse level and pay more attention to the discourses that are key to the explanatory semantics? Inspired by previous work \cite{li-etal-2016-discourse}, we propose a top attention-based discourse-level salient network to focus on the key discourses in terms of explanatory semantics.
	
	In summary, the contributions of this paper are as follows: 
	\begin{itemize}
		\item We design a \textbf{P}yramid \textbf{S}alient-\textbf{A}ware \textbf{N}etwork (PSAN) to detect causal explanations of messages which can effectively learn the pivotal relations between keywords at word level and further filter the key information at discourse level in terms of explanatory semantics.
		
		\item  PSAN can assist in causal explanation detection via capturing the salient semantics of discourses contained in their keywords with a bottom graph-based word-level salient network. Furthermore, PSAN can modify the dominance of discourses via a top attention-based discourse-level salient network to enhance explanatory semantics of messages.
		
		\item Experimental results on the open-accessed commonly used datasets show that our model achieves the best performance. Our experiments also prove the effectiveness of each module.
		
	\end{itemize}
	
	\section{Related Works}
	\textbf{Causal Semantic Detection:} Recently, causality detection which detects specific causes and effects and the relations between them has received more attention, such as the researches proposed by Li \cite{li2019knowledge}, Zhang \cite{zhang2017position}, Bekoulis \cite{bekoulis2018adversarial}, Do \cite{do2011minimally}, Riaz \cite{riaz2014depth}, Dunietz \cite{dunietz2017automatically} and Sharp \cite{sharp2016creating}.
	Specifically, to extract the causal explanation semantics from the messages in a general level, some researches capture the causal semantics in messages from the perspective of discourse structure, such as capturing counterfactual conditionals from a social message with the PDTB discourse relation parsing \cite{son2017recognizing}, a pre-trained model with Rhetorical Structure Theory Discourse Treebank (RSTDT) for exploiting discourse structures on movie reviews \cite{ji2017neural}, and a two-step interactive hierarchical Bi-LSTM framework \cite{xia-ding-2019-emotion} to extract emotion-cause pair in messages. Furthermore, Son \shortcite{son2018causal} defines the causal explanation analysis task (CEA) to extract causal explanatory semantics in messages and annotates a dataset for other downstream tasks. In this paper, we focus on causal explanation detection (CED) which is the fundamental and important subtask of CEA.
	
	\textbf{Syntactic Dependency with Graph Network}: Syntactic dependency is a vital linguistic feature for natural language processing (NLP). There are some researches employ syntactic dependency such as retrieving question answering passage assisted with syntactic dependency \cite{cui2005question}, mining opinion with syntactic dependency \cite{wu-etal-2009-phrase} and so on. For tasks related to causal semantics extraction from relevant texts, dependency syntactic information may evoke causal relations between discourse units in text \cite{gao-etal-2019-modeling}.
	And recently, there are some researches \cite{marcheggiani-titov-2017-encoding,zhang-etal-2018-graph} convert the syntactic dependency into a graph with graph convolutional network (GCN) \cite{kipf2016semi} to effectively capture the syntactic dependency semantics between words in context, such as a semantic role model with GCN \cite{marcheggiani-titov-2017-encoding}, a GCN-based model assisted with a syntactic dependency to improving relation extraction \cite{zhang-etal-2018-graph}. In this paper, we capture the salient explanatory semantics based on the syntactic-centric graph.
	
	\textbf{Self-attention Mechanism:} Self-attention has been introduced to machine translation by Vaswani \cite{Vaswani2017AttentionIA} for capturing global dependencies between input and output and achieved state-of-the-art performance. For language understanding tasks, Shen \cite{shen2018disan} exploits self-attention to learn long-range dependencies. Tan \cite{tan2018deep} applies self-attention for semantic role labeling task and
	achieves state-of-the-art results. In this paper, we utilize a multi-head self-attention encoder to capture the representation of words.
	
	\section{Methodology}
\begin{figure*}
		\centering
		\includegraphics*[width=0.85\textwidth,height=0.36\textheight]{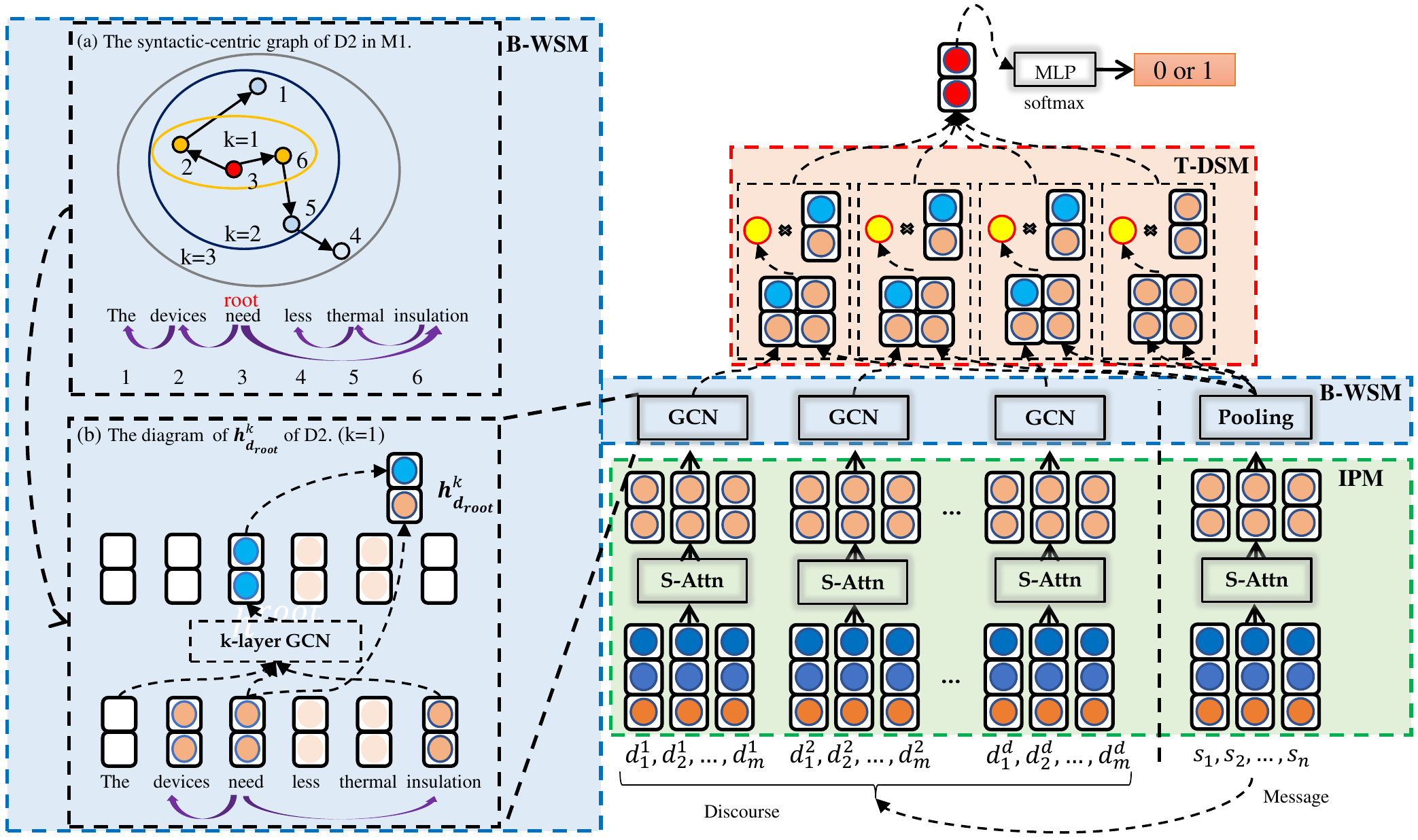}
		\caption{The structure of the pyramid salient-aware network (PSAN). The left side is the detail of the bottom word-level salient-aware module (B-WSM), the top of right side is the top discourse-level salient-aware module (T-DSM) and the bottom of right side is the input processing module (IPM).} \label{fig2}
	\end{figure*}

	The architecture of our proposed model is illustrated in Figure \ref{fig2}. In this paper, the Pyramid Salient-Aware Network (PSAN) primarily involves the following three components: (i) \textbf{input processing module (IPM)}, which processes and encodes the input message and its discourses via self-attention module; (ii) \textbf{bottom word-level salient-aware module (B-WSM)}, which captures the salient semantics of discourses contained in their keywords based on the syntactic-centric graph; (iii) \textbf{top discourse-level salient-aware module (T-DSM)}, which modifies the dominance of different discourse based on the message-level constraint in terms of explanatory semantic via an attention mechanism, and obtain the final causal explanatory representation of input message $m$.
	
	\subsection{Input Processing Module}
	\label{sec:IP}
	In this component, we split the input message $m$ into discourses $d$. Specially, we utilize the self-attention encoder to encode input messages and their corresponding discourses. 
	
	\subsubsection{Discourse Extraction}
	As shown in Figure \ref{fig1}, we split the message into discourses with the same segmentation methods as Son \shortcite{son2018causal} based on semantic coherence. In detail, first, we regard (‘,’), (‘.’), (‘!’), (‘?’) tags and periods as discourse makers. Next, we also extract the discourse connectives set from PDTB2 as discourse makers. Specifically, we remove some simple connectives (e.g. I like running \textbf{and} basketball) from extracted discourse marks. Finally, we divide messages into discourses by the discourse makers. 
	
	\subsubsection{Embedding Layer}
	For the input message $s=\{s_1,...,s_n\}$ and discourse $d=\{d_1^d,...,d_m^d\}$ separated from $s$, we lookup embedding vector of each word $s_n$ ($d_m^d$) as $\bm{s_n}$ ($\bm{d_m^d}$) from the pre-trained embedding. Finally, we obtain the word representation sequence $\bm{s}=\{\bm{s_1},...,\bm{s_n}\}$ of message $s$ and $\bm{d}=\{\bm{d_1^d},...,\bm{d_m^d}\}$ of discourse $d$ corresponding to $s$.

	\subsubsection{Word Encoding}
	Inspired by the application of self-attention  to multiple tasks \cite{tan2018deep,cao-etal-2018-adversarial},  we exploit multi-head self-attention encoder to encode input words. The scaled dot-product attention can be described as follows:
	\begin{equation}
	(\bm{Q}, \bm{K}, \bm{V})=\operatorname{softmax}\left(\frac{\bm{Q} \bm{K}^{T}}{\sqrt{\bm{d}}}\right) \bm{V}
	\end{equation}
	where $\bm{Q} \in \mathbb{R}^{N \times 2dim_{h}}$, $\bm{K} \in \mathbb{R}^{N \times 2 dim_{h}}$ and  $\bm{V} \in \mathbb{R}^{N \times 2 dim_{h}}$  are query matrices, keys matrices and value matrices, respectively. In our setting, $\bm{Q} = \bm{K} = \bm{V} = \bm{s}$ for encoding sentence, and $\bm{Q} = \bm{K} = \bm{V} = \bm{d}$ for encoding discourse.
	
	Multi-head attention first projects the queries, keys, and values $h$ times by using different linear projections. The results of attention are concatenated and once again projected to get the final representation.
	The formulas are as following:
	\begin{equation}
	\begin{aligned}
	head_{i}&=\text { Attention }\left(\mathbf{Q} \mathbf{W}_{i}^{Q}, \mathbf{K} \mathbf{W}_{i}^{K}, \mathbf{V} \mathbf{W}_{i}^{V}\right)  
	\end{aligned}
	\end{equation}
	\begin{equation}
	\begin{aligned}
	\mathbf{H}^{\prime}&=\left(h e a d_{i} \oplus \ldots \oplus h e a d_{h}\right) \mathbf{W}_{o}
	\end{aligned}
	\end{equation}
	where, $\mathbf{W}_{i}^{Q} \in \mathbb{R}^{2 dim_{h} \times dim_{k}}$, $\mathbf{W}_{i}^{K} \in \mathbb{R}^{2 dim_{h} \times dim_{k}}$, $\mathbf{W}_{i}^{V} \in \mathbb{R}^{2 dim_{h} \times dim_{k}}$ and $\mathbf{W}_{o} \in \mathbb{R}^{2 dim_{h} \times 2 dim_{h}}$ are projection parameters and $dim_{k}=2 dim_{h} / h$. And the output is the encoded message $\bm{H}_{S}^{ed}=\{\bm{h}_{s_1}^{ed},...,\bm{h}_{s_n}^{ed}\}$ and discourse $\bm{H}_{D^d}^{ed}=\{\bm{h}_{d^d_1}^{ed},...,\bm{h}_{d^d_m}^{ed}\}$.
	
	\subsection{Bottom Word-Level Salient-Aware Module}
	\label{sec:WSN}
	In this component, we aim to capture the salient semantics of discourses contained in their keywords based on syntactic-centric graphs. For each discourse, first, it extracts the syntactic dependency and constructs the syntactic-centric graph. Second, it collects the keywords and their inter-relations to capture the discourse-level salient semantic based on the syntactic-centric graph. 
	
	\subsubsection{Syntactic-Centric Graph Construction}
	We construct a syntactic-centric graph of each discourse based on syntactic dependency to assist in capturing the semantics of discourses. We utilize Stanford CoreNLP tool\footnote{https://stanfordnlp.github.io/CoreNLP/} to extract the syntactic dependency of each discourse and convert them into syntactic-centric graphs. Specifically, in the syntactic-centric graph, the nodes represent words, and the edges represent whether there is a dependency relation between two words or not. As shown in the subplot (a) of Figure \ref{fig2}, \emph{need} is the root word in the syntactic dependency of \emph{"the devices need less thermal insulation"} (D2 in S1), and words which are syntactically dependent on each other are connected with solid lines.
	\subsubsection{Keywords Collection and Salient Semantic Extraction}
	\label{sec:SIE}
	For each discourse, we collect the keywords based on the syntactic-centric graph and capture the salient semantic based on the syntactic-centric graph from its keywords. Firstly, as illustrated in section \ref{sec:INTRO}, we combine the root word and the affiliated words that connected with the root word in $k$ hops as the keywords. For example, as shown in Figure \ref{fig2}, when $k=1$, the keywords are \emph{\{need, devices and insulation\}}, and the keywords are \emph{\{need, devices, insulation, the and thermal\}} when $k=2$. Secondly, inspired by previous works, we utilize $k$-layer graph convolutional network (GCN) \cite{kipf2016semi} to encode the $k$ hops connected keywords based on the syntactic-centric graph. For example, when $k=1$, we encode 1-hop keywords with 1-layer GCN to capture the salient semantic. Specifically, we can capture different degrees of salient semantics by changing the value of $k$. However, it is not the larger the value of $k$, the deeper the salient semantics are captured. Conversely, the larger the $k$, the more noises are likely to be introduced. For example, when $k=1$, \emph{need}, \emph{devices} and \emph{insulation} are enough to express the salient semantic of D2 (working at these temperatures need less insulation). Finally, we select the representation of the root word in the final layer as the discourse-level representation which contains the salient semantic.
	
	The graph convolutional network (GCN) \cite{kipf2016semi} is a generalization of convolutional neural network \cite{lecun1998gradient} for encoding graphs. In detail, given a syntactic-centric graph with $v$ nodes, we utilize an $v \times v$ adjacency matrix $\bm{A}$, where $A_{ij} = 1$ if there is an edge between node $i$ and node $j$. In each layer of GCN, for each node, the input is the output $\bm{h}_i^{k-1}$ of the previous layer (the input of the first layer is the original encoded input words and features) and the output of node $i$ at $k$-th layer is $\bm{h}_i^k$, the formula is as following:
	\begin{equation}
	\bm{h}_{i}^{k}=\sigma\left(\sum_{j=1}^{v} A_{i j} W^{k} \bm{h}_{j}^{k-1}+b^{k}\right)
	\end{equation}
	where $W^k$ is the matrice of linear transformation, $b^k$ is a bias term and $\sigma$ is a nonlinear function. 
	
	However, naively applying the graph convolution operation in Equation (3) could lead to node representations with drastically different magnitudes because the degree of a token varies a lot. This issue may cause the information in $h_i^{k-1}$ is never carried over to $h_i^k$ because nodes never connect to themselves in a dependency graph  \cite{zhang-etal-2018-graph}. In order to resolve the issue that the information in $h_i^{k-1}$ may be never carried over to $h_i^k$ due to the disconnection between nodes in a dependency graph, we utilize the method raised by Zhang \shortcite{zhang-etal-2018-graph} which normalizes the activations in the GCN, and adds self-loops to each node in graph:
	\begin{equation}
	\bm{h}_{i}^{k}=\sigma\left(\sum_{j=1}^{v} \tilde{A}_{i j} W^{k} \bm{h}_{j}^{k-1} / d_{i}+b^{k}\right),
	\end{equation}
	where $\tilde{\mathbf{A}}=\mathbf{A}+\mathbf{I}$, $\mathbf{I}$ is the $v \times v$ identify matrix and $d_{i}=\sum_{j=1}^{v} \tilde{A}_{i j}$ is the degree of word $i$ in graph. 
	
	Finally, We select the representation $\bm{h}^{k}_{d_{root}}$ of the root word in final layer GCN as the salient representation of $d$-th discourse in message $s$. For example, as shown in the subplot (b) of Figure \ref{fig2}, we choose the representation of \emph{need} in the final layer as the salient representation of the discourse \emph{"the devices need less thermal insulation"}. 
	
	\subsection{Top Discourse-Level Salient-Aware Module}
	\label{sec:DSN}
	How to make better use of the relation between discourse and extract the message-level salient semantic? We modify the dominance of different discourse based on the message-level constraint in terms of explanatory semantic via an attention mechanism. First, we extract the global semantic of message $s$ which contains its causal explanatory tendency. Next, we modify the dominance of different discourse based on global semantic. Finally, we combine the modified representation to obtain the final causal explanatory representation of input message $s$.
	
	\subsubsection{Global Semantic Extraction}
	Inspired by previous research \cite{son2018causal}, the average encoded word representation of all the words in message can represent its overall semantic simply and effectively. We utilize the average pooling on the encoded representation $\bm{H}_{S}^{ed}$ of message $s$ to obtain the global representation which contains the global semantic of its causal explanatory tendency. The formula is as following:
	
	\begin{equation}
	\bm{h}_{s}^{glo}=\sum_{\bm{h}^{ed}_{s} \in \bm{H}_{S}^{ed}} \bm{h}^{ed}_{s} / n,
	\end{equation}
	where $\bm{h}_{s}^{glo}$ is the global representation of message $s$ via average pooling operation and $n$ is the number of words.
	
	\subsubsection{Dominance Modification}
	We modify the dominance of different discourse based on the global semantic which contains its causal explanatory tendency via an attention mechanism. In detail, after obtaining the global representation $\bm{h}_{s}^{glo}$, we modify the salient representation $\bm{h}^{k}_{d_{root}}$ of discourses $d$ constrained with  $\bm{h}_{s}^{glo}$. Finally, we obtain final causal representation $\bm{h}^{caul}_{s}$ of message $s$ via attention mechanism:
	
	\begin{equation}
	\alpha_{ss} = \bm{h}_{s}^{glo} \bm{W}_{f} (\bm{h}_{s}^{glo})^T
	\end{equation}
	
	\begin{equation}
	\alpha_{sd} = \bm{h}_{s}^{glo} \bm{W}_{f} (\bm{h}_{d_{root}}^k)^T 
	\end{equation}
	
	\begin{equation}
	\begin{bmatrix}
	\alpha_{ss}^{'}, \cdots, \alpha_{sd}^{'}
	\end{bmatrix}
	= softmax([\alpha_{ss}, ..., \alpha_{sd}])\\ 
	\end{equation}
	
	\begin{equation}
	\bm{h}^{caul}_{s} = \alpha_{ss}^{'} \bm{h}_{s}^{glo} +...+\alpha_{sd}^{'} \bm{h}_{d_{root}}^k, \\
	\end{equation}
	where the $\bm{W}_{f}$ is matrice of linear transformation, $\alpha_{ss}^{'}$, $\alpha_{sd}^{'}$ are the attention weight. Finally, we mapping $\bm{h}^{caul}_{s}$ into a binary vector and get the output via a softmax operation.  
	
	\section{Experiment}
	\textbf{Dataset} We mainly evaluate our model on a unique dataset devoted to causal explanation analysis released by Son \shortcite{son2018causal}. This dataset contains 3,268 messages consist of 1598 positive messages that contain a causal explanation and 1670 negative sentences randomly selected. Annotators annotate which messages contain causal explanations and which text spans are causal explanations (a discourse with a tendency to interpret something). We utilize the same 80\% of the dataset for training, 10\% for tuning, and 10\% for evaluating as Son \shortcite{son2018causal}. Additionally, to further prove the effectiveness of our proposed model, we regard sentences with causal semantic discourse relations in PDTB2 and sentences containing causal span pairs in BECauSE Corpus 2.0 \cite{dunietz-etal-2017-corpus} as supplemental messages with causal explanations to evaluate our model. In this paper, PDTB-CED and BECauSE-CED are used to represent the two supplementary datasets respectively.
	
	\textbf{Parameter Settings} We set the length of the sentence and discourse as 100 and 30 respectively. We set the batch size as 5 and the dimension of the output in each GCN layer as 50. Additionally, we utilize the 50-dimension word vector pre-trained with Glove. For optimization, we utilize Adam \cite{kingma2014adam} with 0.001 learning rate. We set the maximum training epoch as 100 and adopt an early stop strategy based on the performance of the development set. All the results of different compared and ablated models are the average result of five independent experiments.
	
	\textbf{Compared Models} We compare our proposed model with feature-based  and neural-based model: (1) \textbf{Lin et al. \shortcite{lin2014pdtb}}: an end-to-end discourse relation parser on PDTB, (2) \textbf{Linear SVM}: a linear designed feature based SVM classifier, (3) \textbf{RBF SVM}: a complex designed feature based SVM classifier, (4) \textbf{Random Forest}: a random forest classifier which relies on designed features, (5) \textbf{Son et al. \shortcite{son2018causal}}: a hierarchical LSTM sequence model which is designed specifically for CEA. (6) \textbf{H-BiLSTM + BERT}\footnote{https://github.com/huggingface/transformers}\footnote{BERT can not be applied to the feature-based model suitably, so we deploy BERT on the latest neural network model to make the comparison to prove the effectiveness of our proposed model.}: a fine-tuned language model (BERT) which has been shown to improve the performance in some other classification tasks based on (5), (7) \textbf{H-Atten.}: a well-used Bi-LSTM model that captures hierarchical key information based on hierarchical attention mechanism, (8) \textbf{Our model}: our proposed pyramid salient-aware network (PSAN). Furthermore, we evaluate the performance of the model (5), (7), and (8) on the supplemental dataset to prove the effectiveness of our proposed model. Additionally, we design different ablation experiments to demonstrate the effectiveness of the bottom word-level salient-aware module (B-WSM), top discourse-level salient-aware module (T-DSM), and the influence of different depths in the syntactic-centric graph.
	
	\subsection{Main Results}
	
	\begin{table}[h]
	\centering
	\label{tab1}
	\begin{tabular}{c|c|c|c}
		\hline
		\textbf{Model} & \textbf{F1} & \textbf{F1} & \textbf{F1} \\ \hline
		& \textbf{Facebook} & \textbf{PDTB-CED} & \textbf{BEcuasE-CED}  \\ \hline
	    Lin et al. \shortcite{lin2014pdtb}    & 63.8   & - & -       \\ \hline
		Linear SVM \cite{son2018causal}     & 79.1    & - & -        \\ \hline
		RBF SVM \cite{son2018causal}        & 77.7   & - & -        \\ \hline
		Random Forest \cite{son2018causal}  & 77.1    & - & -        \\ \hline
		Son et al. \shortcite{son2018causal}      & 75.8    & 63.6  & 69.6          \\ \hline
		H-Atten.      & 80.9     & 70.6    & 76.5   \\ \hline
		H-BiLSTM + BERT      & 85.0  & - & -         \\ \hline
		\textbf{Our model}     & \textbf{86.8}  & \textbf{76.6}   & \textbf{81.7}  \\ \hline
		
		\end{tabular}
		\caption{Comparisons of the state-of-the-art methods on causal explanation detection.}
	\end{table}
	
	Table 1 shows the comparison results on the Facebook dataset and two supplementary datasets. From the results, we have the following observations.
	
	(1) Comparing with the current best feature-based and neural-based models on CED: \textbf{Lin et al. \shortcite{lin2014pdtb}}, \textbf{Linear SVM} and \textbf{Son et al. \shortcite{son2018causal}}, \textbf{our model} improves the performance by 23.0, 7.7 and 11.0 points on F1, respectively. It illustrates that the pyramid salient-aware network (PSAN) can effectively extract and incorporate the word-level key relation and discourse-level key information in terms of explanatory semantics to detect causal explanation. Furthermore, comparing with the well-used hierarchical key information captured model (\textbf{H-Atten.}), \textbf{our model} improves the performance by 5.9 points on F1. This confirms the statement in section \ref{sec:INTRO} that directly employing the relation between words with syntactic structure is more effective than the implicit learning.
	
	(2) Comparing the \textbf{Son et al. \shortcite{son2018causal}} with pre-trained language model (\textbf{H-BiLSTM+BERT}), there is 9.2 points improvement on F1. It illustrates that the pre-trained language model (LM) can capture some causal explanatory semantics with the large-scale corpus. Furthermore, \textbf{our model} can further improve performance by 1.8 points compared with \textbf{H-BiLSTM+BERT}. We believe the reason is that the LM is pre-trained with large-scale regular sentences that do not contain causal semantics only, which is not specifically suitable for CED compared to the proposed model for explanatory semantic. Furthermore, the performance of \textbf{H-Atten.} is better than \textbf{Son et al. \shortcite{son2018causal}} which indicates focusing on salient keywords and key discourses helps understand explanatory semantics.
	
	(3) It is worth noting that, regardless of our proposed model, comparing the \textbf{Linear SVM} with  \textbf{Son et al. \shortcite{son2018causal}}, the simple feature classifier is better than the simple deep learning model for CED on the Facebook dataset. However, when combining the syntactic-centric features with deep learning, we could achieve a significant improvement. In other words, our model can effectively combine the \emph{interpretable information} of the feature-based model with the \emph{deep understanding} of the deep learning model.
	
	(4) To further prove the effectiveness of the proposed model, we evaluate \textbf{our model} on supplemental messages with causal semantics in other datasets (PDTB-CED and BEcausE-CED). As shown in Table 1, the results show that the proposed model performs significantly better than the \textbf{Son et al. \shortcite{son2018causal}} and \textbf{H-Atten.} on the other two datasets\footnote{We obtain the performance with the publicly released code by Son et al. \shortcite{son2018causal}. The supplementary datasets are not specifically suitable for this task, and the architectural details of designed feature-based models are not public, so we only compare the performance of the latest model to prove the effectiveness of our proposed model.}. It further demonstrates the effectiveness of our proposed model.
	
	(5) Moreover, \textbf{our model} is twice as fast as the \textbf{Son et al. \shortcite{son2018causal}} during training because of the computation of self-attention and GCN is parallel. It illustrates that our model can consume less time and achieve significant improvement in causal explanation detection. Moreover, compared with the feature-based models, the neural-based models rely less on artificial design features.
	
	\subsection{Effectiveness of Bottom Word-Level Salient-Aware Module (B-WSM)}
	\begin{table}[h]
		\centering
		\label{tab3}
		\begin{tabular}{c|cccccc}
			\hline
			\textbf{Dataset}      & \multicolumn{2}{c}{\textbf{Facebook}}   & \multicolumn{2}{c}{\textbf{PDTB-CED}}   & \multicolumn{2}{c}{\textbf{BEcausE-CED}}   \\ \hline
			\textbf{Model}        & \textbf{F1}   & \bm{$\nabla$}  & \textbf{F1}   & \bm{$\nabla$}  & \textbf{F1}   & \bm{$\nabla$}\\ \hline
			\textbf{\textbf{our model}}   & \textbf{86.8} & - & \textbf{76.6}  & -  & \textbf{81.7}  & - \\ \hline
			w/o B-WSM + root  & 80.1     &  -6.7          & 69.9     &  -6.7  & 75.8     &  -5.9  \\ \hline
			w/o B-WSM + ave   & 84.7     &  -2.1          & 74.4     &  -2.2  & 79.8     &  -1.9  \\ \hline
		\end{tabular}
		\caption{Effectiveness of B-WSM. (\textbf{w/o} B-WSM denotes the models without B-WSM. $\bm{+}$ denotes repalcing the B-WSM with the module after $\bm{+}$.
			\textbf{root} denotes using the encoded representation of the root word in each discourse to represent it. \textbf{ave} denotes using the average encoded representation of words in discourse to represent it.)}
	\end{table}
	
	Table 2 tries to show the effectiveness of the salient information contained in the keywords of each discourse captured via the proposed B-WSM for causal explanation detection (\ref{sec:WSN}). The results illustrate B-WSM can effectively capture the salient information which contains the most causal explanatory semantics. It is worth noting that when using the average encoded-word representation to represent each discourse (\textbf{w/o B-WSM + ave}), the model also achieves acceptable performance. This confirms the conclusion from Son \shortcite{son2018causal} that the average word representation at word level contains certain causal explanatory semantic. Furthermore, only the root word of each discourse also contains some causal semantics (\textbf{w/o B-WSM + root}) which proves the effectiveness of capturing salient information via syntactic dependency from the keywords. 
	
	\subsection{Effectiveness of Top Discourse-Level Salient-Aware Module (T-DSM)}
	\begin{table}[h]
		\centering
		\label{tab4}
		\begin{tabular}{c|cccccc}
			\hline
			\textbf{Dataset}      & \multicolumn{2}{c}{\textbf{Facebook}}   & \multicolumn{2}{c}{\textbf{PDTB-CED}}   & \multicolumn{2}{c}{\textbf{BEcausE-CED}}   \\ \hline
			\textbf{Model}        & \textbf{F1}   & \bm{$\nabla$}  & \textbf{F1}   & \bm{$\nabla$}  & \textbf{F1}   & \bm{$\nabla$}\\ \hline
			\textbf{our model} & \textbf{86.8}          & -          & \textbf{76.6}     &  -   & \textbf{81.7}     &  - \\ \hline
			w/o T-DSM + seq D  & 83.8      &  -3.0          & 72.9      &  -3.7    & 78.1      &  -3.6  \\ \hline
			w/o T-DSM + ave S/D & 84.0     & -2.8     & 73.5  & -3.1   & 77.8      &  -3.9 \\ \hline
		\end{tabular}
		\caption{Effectiveness of T-DSM. (\textbf{w/o} T-DSM denotes models without T-DSM. $\bm{+}$ denotes replacing the T-DSM with the module after $\bm{+}$. \textbf{seq D} denotes mapping the representation of discourses via a sequence LSTM to represent the whole message. \textbf{ave S/D} denotes using the average encoded representation of words in message and its discourses to represent the whole message.)}
	\end{table}
	
	Table 3 tries to show the effectiveness of the salient information of the key discourses modified and incorporated via T-DSM for causal explanation detection (\ref{sec:DSN}). The results compared with \textbf{w/o T-DSM + seq D} illustrate our T-DSM can effectively modify the dominance of different discourses based on the global semantic constraint via an attention mechanism to enhance the causal explanatory semantic. Specifically, the results of \textbf{w/o T-DSM + ave S/D} show that both discourse-level representation and global representation contain efficient causal explanatory semantics, which further proves the effectiveness of the proposed T-DSM.
	
	\subsection{Comparisons of Different Depths of Syntactic-Centric Semantic}
	
	\begin{figure}[h] 
		\centering
		\includegraphics*[width=0.45\textwidth,height=0.23\textheight]{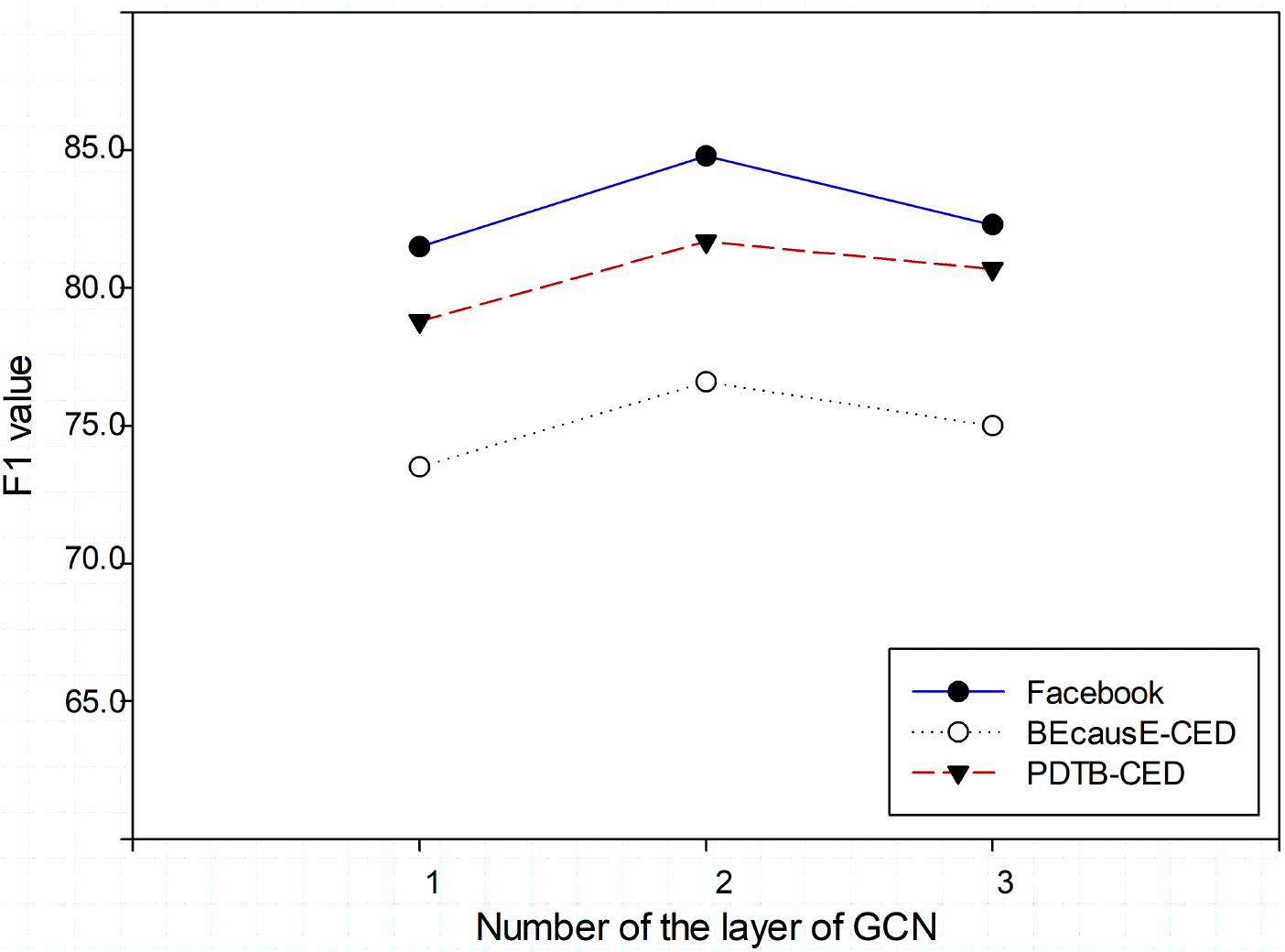}
		\caption{Comparisons of different number of GCN layers.} \label{fig3}
	\end{figure}
	
	To demonstrate the influence of the causal explanatory semantics contained in the syntactic-centric graph with different depths, we further compare the performance of our proposed model with a different number of GCN layers. As shown in Figure \ref{fig3}, when the number of GCN layers is 2, the most efficient syntactic-centric information can be captured for causal explanation detection. 
	
	\subsection{Error Analysis}
	As shown in Figure \ref{fig4}, we find the two main difficulties in this task:
	
	\begin{figure}[h]
		\centering
		\includegraphics*[width=0.42\textwidth,height=0.12\textheight]{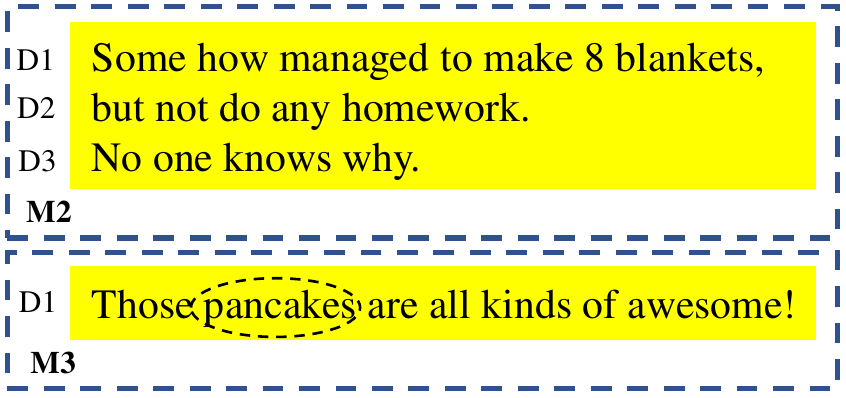}
		\caption{Predictions of the proposed model.} \label{fig4}
		\vspace{-8pt}
	\end{figure}
	
	(1) \textbf{Emotional tendency} The same expression can convey different semantic under different emotional tendencies, especially in this kind of colloquial expressions. As M2 shown in Figure \ref{fig4}, \emph{make 8 blankets} expresses \emph{anger} over \emph{not do any homework}, and our model wrongly predicts the \emph{make 8 blankets} is the reason for \emph{not do any homework}.
	
	(2) \textbf{Excessive semantic parsing} Excessive parsing of causal intent by the model will lead to identifying messages that do not contain causal explanations as containing. As shown in Figure \ref{fig4}, M3 means pancakes are awesome, but the model overstates the reason for \emph{awesome} is a pancake.
	
	\section{Conclusion}
	In this paper, we devise a pyramid salient-aware network (PSAN) to detect causal explanations in messages. PSAN can effectively learn the key relation between words at the word level and further filter out the key information at the discourse level in terms of explanatory semantics. Specifically, we propose a bottom word-level salient-aware module to capture the salient semantics of discourses contained in their keywords based on a the syntactic-centric graph. We also propose a top discourse-level salient-aware module to modify the dominance of different discourses in terms of global explanatory semantic constraint via an attention mechanism. Experimental results on the open-accessed commonly used datasets show that our model achieves the best performance.
	
	\section*{Acknowledgements}
	This work is supported by the Natural Key RD Program of China (No.2018YFB1005100), the National Natural Science Foundation of China (No.61533018, No.61922085, No.61976211, No.61806201) and the Key Research Program of the Chinese Academy of Sciences (Grant NO. ZDBS-SSW-JSC006). This work is also supported by Beijing Academy of Artificial Intelligence (BAAI2019QN0301), CCF-Tencent Open Research Fund and independent research project of National Laboratory of Pattern Recognition.
	
	\newpage
	
	\bibliographystyle{ccl}
	\bibliography{ccl2020-en}
	
\end{document}